%% file: main.tex
\documentclass[10pt,twocolumn,letterpaper]{article}

\usepackage{cvpr}      % To produce the REVIEW version
\input{preamble}

\definecolor{cvprblue}{rgb}{0.21,0.49,0.74}
\usepackage[pagebackref,breaklinks,colorlinks,citecolor=cvprblue]{hyperref}

\def\paperID{810} % *** Enter the Paper ID here
\def\confName{CVPR}
\def\confYear{2025}

\title{EasyCraft: A Robust and Efficient Framework for Automatic Avatar Crafting}
\author{
Suzhen Wang\textsuperscript{1},\enspace
Weijie Chen\textsuperscript{1},\enspace
Wei Zhang\textsuperscript{1},\enspace
Minda Zhao\textsuperscript{1},\enspace
Lincheng Li\textsuperscript{1}\footnotemark[1],\enspace \\
Rongsheng Zhang\textsuperscript{1},\enspace
Zhipeng Hu\textsuperscript{1},\enspace
Xin Yu\textsuperscript{2}\\
\textsuperscript{1}Netease Fuxi AI Lab,\enspace
\textsuperscript{2}The University of Queensland\\
{\tt\small \{wangsuzhen, chenweijie05, zhangwei05, zhaominda01, lilincheng\}@corp.netease.com,} \\
{\tt\small \{zhangrongsheng, zphu\}@corp.netease.com,} {\tt\small xin.yu@uq.edu.au}
}
\begin{document}
% \maketitle

\twocolumn[{%
\renewcommand\twocolumn[1][]{#1}%
\maketitle

\begin{center}
    \centering
    \captionsetup{type=figure}
    \includegraphics[width=1.0\textwidth]{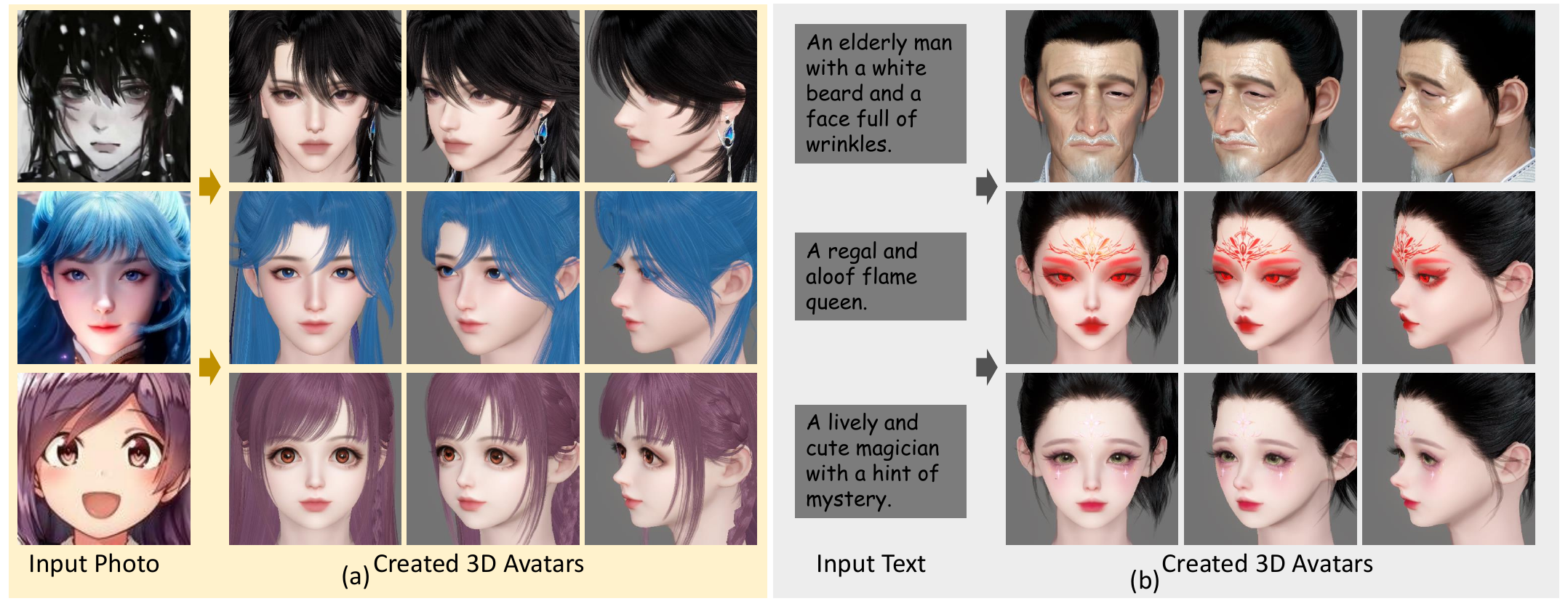}
    \captionof{figure}{Illustration of EasyCraft. The proposed method can achieve both (a) photo-based avatar auto-creation using any style of photo input, and (b) text-based avatar auto-creation from text descriptions.}
    \label{fig:teaser}
\end{center}%
}]

\footnotetext[1]{* Corresponding Author.}
\begin{abstract}
Character customization, or 'face crafting,' is a vital feature in role-playing games (RPGs), enhancing player engagement by enabling the creation of personalized avatars. Existing automated methods often struggle with generalizability across diverse game engines due to their reliance on the intermediate constraints of specific image domain and typically support only one type of input, either text or image. To overcome these challenges, we introduce EasyCraft, an innovative end-to-end feedforward framework that automates character crafting by uniquely supporting both text and image inputs.
Our approach employs a translator capable of converting facial images of any style into crafting parameters. We first establish a unified feature distribution in the translator's image encoder through self-supervised learning on a large-scale dataset, enabling photos of any style to be embedded into a unified feature representation. Subsequently, we map this unified feature distribution to crafting parameters specific to a game engine, a process that can be easily adapted to most game engines and thus enhances EasyCraft's generalizability. By integrating text-to-image techniques with our translator, EasyCraft also facilitates precise, text-based character crafting. EasyCraft's ability to integrate diverse inputs significantly enhances the versatility and accuracy of avatar creation. Extensive experiments on two RPG games demonstrate the effectiveness of our method, achieving state-of-the-art results and facilitating adaptability across various avatar engines.

\end{abstract}

\section{Introduction}

Character customization, often referred to as ``face crafting,'' is a fundamental feature in role-playing games (RPGs) that enables players to create and personalize their in-game avatars. Over time, this feature has evolved to include numerous options for modifying facial structures, hairstyles, makeup, and other aesthetic elements, allowing players to design unique characters. This customization enhances player engagement by fostering a personal connection with their avatars and enriching the overall gaming experience. However, the process can become laborious and time-consuming, especially when players aim to achieve specific appearances, such as resembling a celebrity or embodying a particular style, often requiring extensive manual adjustments. With rapid technological advancements, there is growing interest in automating the character customization process. Yet, automating the generation of face crafting parameters from specific inputs, such as photos or text descriptions, remains challenging due to the significant distribution gap between inputs and desired outputs.

% Character customization, often known as ``face crafting,'' is a fundamental feature in role-playing games (RPGs) that enables players to create and personalize their in-game avatars. Over time, this feature has evolved to include numerous options for modifying facial structures, hairstyles, makeup, and other aesthetic elements, allowing players to design unique characters. This customization enhances player engagement by fostering a personal connection with their avatars and enriching the overall gaming experience. However, the process can become laborious and time-consuming, particularly when players strive to achieve specific appearances, such as resembling a celebrity or embodying a particular style, often requiring extensive manual adjustments. With rapid technological advancements, there is growing interest in automating the character customization process. Yet, automating the generation of face crafting parameters from specific inputs, such as photos or text descriptions, remains challenging due to the significant distribution gap between inputs and desired outputs.

Current methods attempt to leverage semantic constraints between the avatar face image domain and the target domain, utilizing techniques such as segmentation \cite{shi2019face}, perceptual \cite{wang2023swiftavatar, sang2022agileavatar}, and CLIP constraints \cite{zhao2023zero}, as unsupervised signals for face crafting. However, the non-differentiability of parameter controller-based customization engines compels these methods to develop neural renderers that simulate the parameter-to-avatar face image process. This approach enables the transfer of constraints from the image domain to the parameter domain via inversion techniques. Unfortunately, this limits these methods to applying supervision signals only on images of specific styles, necessitating effective off-the-shelf model constraints for the renderer's output image domain (i.e., engine style). Consequently, significant changes in engine style (e.g., realistic, anime, cartoon) or input style (e.g., realistic or cartoon photos, text descriptions) substantially degrade the performance of these methods. This presents a substantial challenge in extending these methods across engines with varying styles. In practical applications, the wide variability in engine styles and user inputs necessitates a more adaptable solution.

% Current methods attempt to leverage semantic constraints between the avatar face image domain and the target domain, such as segmentation \cite{shi2019face}, perceptual \cite{wang2023swiftavatar, sang2022agileavatar}, and CLIP constraints \cite{zhao2023zero}, serving as unsupervised signals for face crafting. However, the non-differentiability of parameter controller-based customization engines leads these methods to develop neural renderers that simulate the parameter-to-avatar face image process. This allows constraints in the image domain to be transferred to the parameter domain via inversion techniques. Unfortunately, this restricts these methods to applying supervision signals only on images of specific styles, requiring effective off-the-shelf model constraints for the renderer's output image domain (i.e., engine style). Consequently, significant changes in engine style (e.g., realistic, anime, cartoon) or input style (e.g., realistic or cartoon photos, text descriptions) sharply reduce the performance of these methods. This presents a substantial challenge in extending these methods across engines with varying styles. In practical applications, the wide variability in engine styles and user inputs necessitates a more adaptable solution for automated face crafting.

To address these challenges, we introduce \textbf{EasyCraft}, an innovative end-to-end feedforward framework for automated character crafting. The core of our method is a translator that converts any facial photo into specific game crafting parameters. Leveraging this translator, we can directly create characters from photos and, by integrating with a text-to-image framework, seamlessly implement character customization through text descriptions.

% To address these challenges, we introduce \textbf{EasyCraft}, an innovative end-to-end feedforward framework for automated character crafting. The core of our method is a translator that converts any facial photo into specific game crafting parameters. Based on this translator, we can directly create characters from photos and, by integrating with a text-to-image framework, easily implement character customization through text descriptions.

Our translator consists of a vision transformer \cite{VIT} encoder that encodes images into features and a parameter generation module that transforms these features into output parameters. By leveraging the game engine, we generate pairs of game-rendered images and crafting parameters to serve as training data for the translator. Naturally, a translator trained solely on engine data faces challenges when dealing with non-game style images due to distribution inconsistencies between the input domains and engine styles. Unlike previous inversion-based methods, which are limited to specific image domains for constraint or alignment, our feedforward approach offers a novel perspective by creating a unified image feature distribution to address this inconsistency. We hypothesize that once the feature distribution encoded by the vision transformer is unified, the translator trained solely on game engine pairs can handle inputs of any style. Motivated by this, we begin by constructing a universal vision transformer encoder for the translator. Specifically, we create a diverse dataset containing facial images in various styles, such as real-life, anime, and engine-generated images. We pretrain our vision transformer encoder on this dataset using a method akin to the Masked Autoencoder (MAE) \cite{he2022masked}. Subsequently, we employ the pre-trained vision transformer encoder as the translator's encoder, keeping its parameters frozen while training only the parameter generation module using the game engine pairs. Since the features produced by the vision transformer and fed into the parameter generation module maintain a consistent distribution across different image types, our translator can process a variety of facial image styles and generate crafting parameters for the game engine, despite being trained only on engine data.

Building on the design of our translator, our approach seamlessly integrates with text-to-image methods, enabling text-based automatic avatar creation. To achieve more precise text-based character crafting, we train a text-to-game-style facial image model using the Stable Diffusion (SD) framework \cite{rombach2022high}. By fine-tuning the model with a small set of annotated text-to-image pairs and leveraging a pre-trained SD module, we can produce images in the style of the game engine while preserving SD's rich semantic capabilities. This approach allows us to fully exploit the diversity of SD, enabling the generation of a wide range of character appearances that fit the same textual description.

% Building on the design of our translator, our approach seamlessly integrates with text-to-image methods, enabling text-based automatic avatar creation. To achieve more precise text-based character crafting, we train a text-to-game-style facial image model utilizing the Stable Diffusion (SD) framework \cite{rombach2022high}. By fine-tuning the model with a small set of annotated text-to-image pairs and leveraging a pre-trained SD module, we can produce images in the style of the game engine while preserving SD's rich semantic generalization capabilities. This approach allows us to fully exploit the diversity of SD, enabling the generation of a wide range of character appearances that fit the same textual description.

In summary, our proposed approach effectively overcomes the limitations of existing methods by providing a seamless, integrated framework for character customization. By incorporating both text and image inputs, \textbf{EasyCraft} enhances the versatility and precision of automatic avatar creation. Since our translator's training relies solely on the game engine without additional supervision, our method can be effortlessly applied to various systems that support character customization. Extensive experiments on two RPG games demonstrate the effectiveness of our approach, achieving state-of-the-art results.

\begin{figure*}
    \centering
    \includegraphics[ width=1\linewidth]{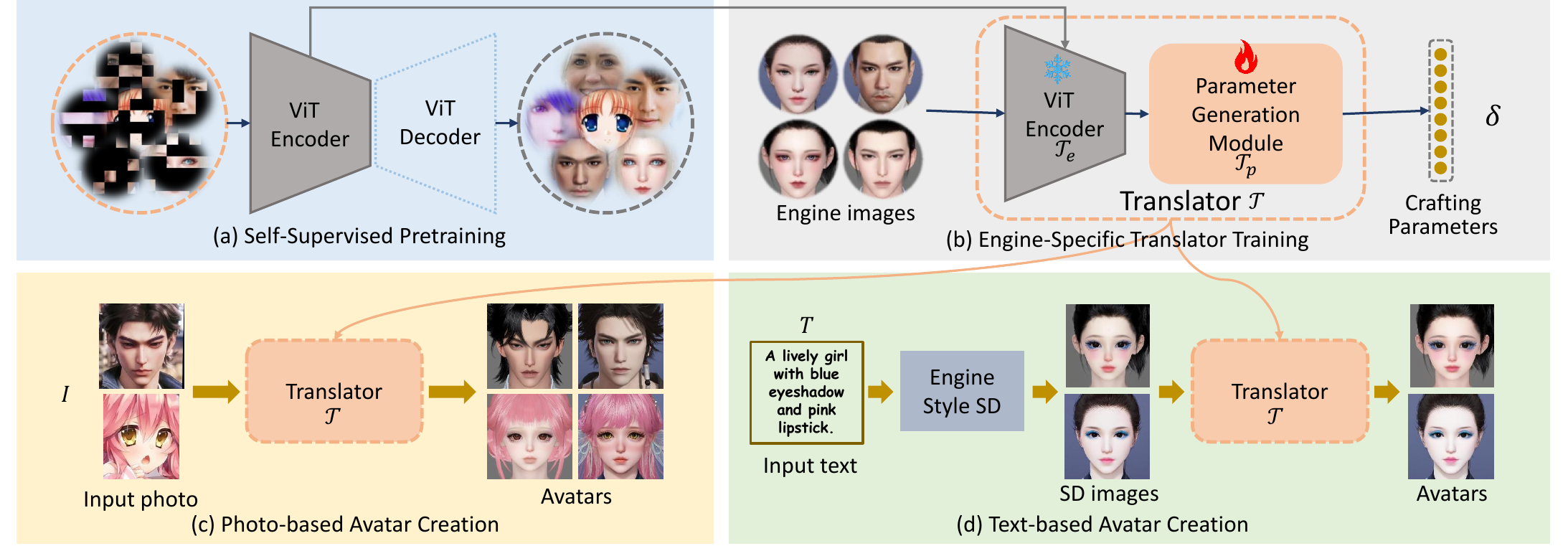}
    %\vspace{-1.5em}
    \caption{Illustration of EasyCraft. (a) We first employ self-supervised learning to develop a universal vision transformer (ViT) encoder using a large-scale dataset containing various styles of photos. (b) We then train an engine-specific translator $\mathcal{T}$ that can convert input images into specific avatar crafting parameters. Our translator consists of a ViT encoder $\mathcal{T}_e$ and a parameter generation module $\mathcal{T}_p$. During this training process, only $\mathcal{T}_p$ is trained, while $\mathcal{T}_e$ is initialized from the pretrained ViT encoder and remains frozen. (c) Once the translator is obtained, we can directly perform photo-based automatic avatar creation. 
    (d) By integrating our SD model, which can generate engine-style photos based on text, our method also facilitates text-based automatic avatar creation.}
    \label{fig:pipeline}
\end{figure*}

\section{Related Work}

\subsection{3D Face Reconstruction}

Various approaches employ computer vision and graphics techniques to generate 3D face models from input photographs \cite{cashman2012shape,tran2019learning,tretschk2023state,ramon2021h3d,sun2023next3d,yi2023generating,tian2023recovering,song2024agilegan3d,hu2024gaussianavatar} or textual descriptions \cite{liu2024headartist,cao2024dreamavatar,liu2024humangaussian,kolotouros2024dreamhuman,han2024headsculpt,lei2024diffusiongan3d,jiang2023avatarcraft}. Early works predominantly leveraged 3D morphable models (3DMM) \cite{deng2019accurate}, the Basel Face Model (BFM) \cite{gerig2018morphable}, or the FLAME model \cite{li2017learning} to accurately fit the texture and geometry of photorealistic 3D face models. Recent advancements have explored the construction of latent 3D face models using Neural Radiance Fields (NeRF) \cite{hwang2023faceclipnerf,sun2023next3d,song2024agilegan3d} or 3D Gaussian models \cite{hu2024gaussianavatar,chen2024monogaussianavatar,ma20243d}. Nonetheless, integrating these techniques into role-playing games presents challenges, as these games require customization through parameter controllers, a process fundamentally distinct from direct 3D modeling.

\subsection{Engine-based Avatar Auto-Creation}
Building supervised datasets for automatic engine-based avatar creation is a labor-intensive and costly process that lacks scalability across different face-crafting engines. Consequently, current methodologies \cite{shi2019face,shi2020neural,shi2020neutral,zhao2023zero,sang2022agileavatar,wang2023swiftavatar,shi2020fast} have increasingly turned to unsupervised approaches. These methods predominantly employ unsupervised constraints utilizing off-the-shelf models, such as segmentation \cite{shi2019face}, perceptual \cite{wang2023swiftavatar, sang2022agileavatar}, and CLIP constraints \cite{zhao2023zero}, to relate the avatar face image generated by the renderer with the input text or image. The efficacy of these constraints is heavily influenced by the diversity of engine styles (e.g., realistic, anime, cartoon) and the variety of input styles (e.g., realistic or cartoon photos, text descriptions). If these models have not been adequately trained on images or input styles specific to a given game style, the effectiveness of applying these constraints diminishes considerably. As a result, these methods are often inadequate for use in other styles of avatar customization engines.

\subsection{Text-to-Image Generation}
Text-to-image generation has seen significant advancements in recent years. Early approaches like AttnGAN \cite{AttnGAN} and StackGAN \cite{zhang2017stackgan, zhang2018stackgan++} introduced attention mechanisms and multi-stage generation processes, which improved image quality and text alignment, though they often struggled with complex scenes and semantic consistency. DALL-E \cite{ramesh2021zero} marked a significant shift by employing an autoregressive transformer architecture to learn joint distributions of text and images. GLIDE \cite{nichol2021glide} further advanced the field by incorporating text conditioning into diffusion models \cite{sohl2015deep, ho2020denoising} and utilizing an upsampler for high-resolution output. Imagen \cite{saharia2022photorealistic} leveraged a pre-trained transformer as a text encoder, substantially enhancing the model's text comprehension capabilities. More recently, Stable Diffusion \cite{rombach2022high} demonstrated unprecedented performance by modeling images in latent space, generating high-quality images from textual descriptions. In this paper, we combine text-to-image generation with our image-to-avatar parameter translator to achieve text-based avatar creation.

\section{Methodology}

% Figure 2 illustrates the pipeline of our proposed method, which unifies photo-based and text-based avatar creation within a single framework. Given an input, either a photo $I$ or a text description $T$  our framework can automatically derive the desired avatar parameters $\delta$.
% The core of our approach is a translator capable of generating the corresponding avatar system parameters from an input image. For photo-based avatar creation, the input image $I$ can be directly converted into avatar parameters using the translator. In the case of text-based avatar creation, we first utilize a Stable Diffusion model to generate image $I_{text}$  in the style of the target avatar system given text description $T$. These images are then transformed into avatar parameters through the translator.

\subsection{Formulation and Overview}

An avatar customization system typically involves three types of parameters: facial structure parameters ($\delta_s \in \mathbb{R}^{D_s}$), makeup texture parameters ($\delta_t \in \mathbb{R}^{D_t}$), and makeup attribute parameters ($\delta_a \in \mathbb{R}^{D_a}$). The parameter $\delta_s$ adjusts the character's facial structure with continuous values for features like eye size, nose width, and mouth position. $\delta_t$ offers discrete choices for makeup textures, such as eyebrow shapes and eyeshadow styles, using one-hot encodings. $\delta_a$ fine-tunes makeup attributes with continuous values, adjusting aspects like eyebrow color and lip gloss brightness. Together, these parameters form the complete crafting parameters $\delta$ for an avatar. Our task involves automatically generating the desired avatar parameters $\delta$ from a photo $I$ or a text description $T$.

Figure 2 presents the pipeline of our proposed method, which integrates photo-based and text-based avatar creation into a unified framework. Initially, our approach utilizes self-supervised learning to develop a universal vision transformer encoder that serves as the image feature extractor. Leveraging this encoder, alongside a specific avatar customization system, we derive a translator capable of converting input images into avatar parameters. This enables direct avatar creation from images of various styles (Sec. \ref{photo-based}). Additionally, by incorporating the SD model for text-to-face-image generation, our method facilitates the creation of avatars from text descriptions (Sec. \ref{text-based}).

\begin{figure*}
\centering
\includegraphics[width=0.98\textwidth]{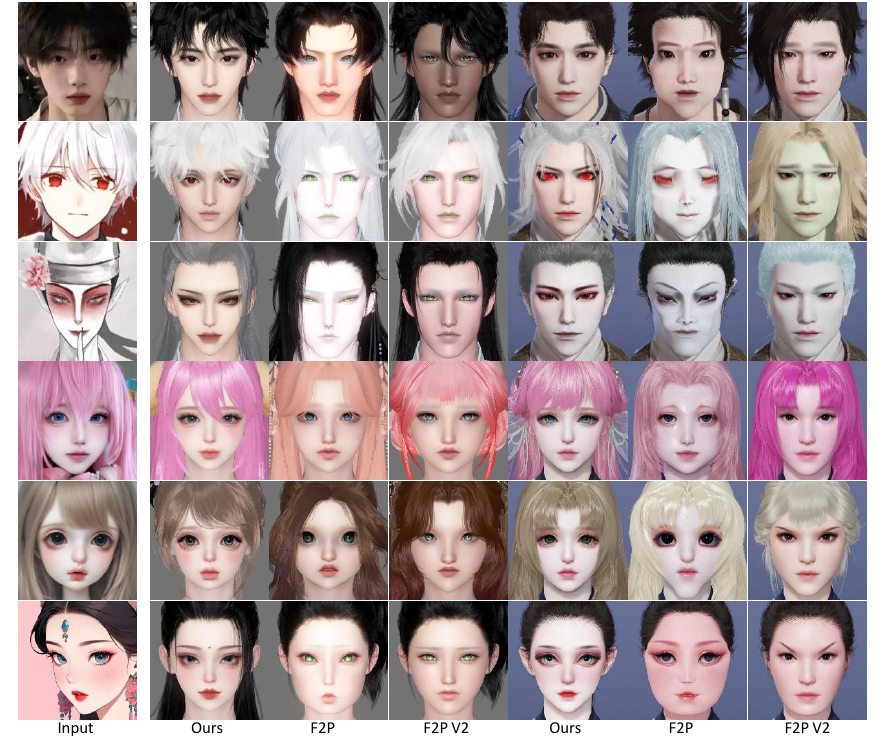}
% {images/ablation_qualitative_head_hightlight.pdf}
\caption{Qualitative comparisons with photo-based methods. For each input image, we show results of three methods on two engines.
}
\label{fig:qualitative_image}
\end{figure*}

\subsection{Photo-base Automatic Avatar Creation}\label{photo-based}

In this paper, we aim to utilize a feedforward approach for automatic facial customization. Leveraging the game engine, it is convenient to obtain paired facial customization parameters and rendered images. With these paired data, we can easily train a translator $\mathcal{T}$ that inputs game-style images and outputs facial customization parameters. However, this translator performs poorly with non-game-style images. To develop a translator that supports image inputs of arbitrary styles based solely on engine data training, we propose a translator structure comprising a vision transformer (ViT) encoder $\mathcal{T}_e$ and a parameter generation module $\mathcal{T}_p$. Additionally, we design a two-stage training pipeline: pre-training the ViT encoder through self-supervised learning to acquire a universal image feature extractor, followed by training the parameter generation module with engine data to convert image features into the target engine's facial customization parameters.

% In this paper, we aim to utilize a feedforward approach for automatic facial customization. Leveraging the game engine, it is convenient to obtain paired facial customization parameters and rendered images. With these paired data, we can easily train a translator $\mathcal{T}$ that inputs game-style images and outputs facial customization parameters. However, this translator evidently performs poorly with non-game-style images. To develop a translator that supports image inputs of arbitrary style based solely on engine data training, we propose a translator structure comprising a vision transformer (VIT) encoder $\mathcal{T}_e$ and a parameter generation module $\mathcal{T}_p$. Additionally, we design a two-stage training pipeline: pre-training the VIT encoder through self-supervised learning to acquire a universal image feature extractor, followed by training the parameter generation module with engine data to convert image features into the target engine's facial customization parameters.

\subsubsection{Universal Feature Extraction}
We adopt the self-supervised learning strategy proposed by Masked Autoencoders (MAE) \cite{he2022masked} to train our ViT encoder $\mathcal{T}_e$. This approach enables the ViT encoder to extract relatively universal facial features from images with diverse styles. Specifically, we first build a facial image dataset that includes various styles, such as real-life, cartoon, anime, and importantly, images from the target engine. We then train the encoder-decoder architecture (see Fig.~\ref{fig:pipeline}(a)) using self-supervised learning on this dataset. Both the encoder and decoder are based on the vision transformer structure, and we append a global [CLS] token in $\mathcal{T}_e$. During training, each image is divided into $16 \times 16$ patches, and 75\% of these patches are randomly masked before being fed into the ViT encoder. The encoder processes the masked images into tokens, and the ViT decoder attempts to reconstruct the original image from these tokens using pixel-wise L2 loss. Through pre-training on datasets with various styles, we obtain a universal $\mathcal{T}_e$ that effectively extracts both the structural and cosmetic characteristics of facial images.

% In this paper, we adopt the self-supervised learning strategy proposed by Masked Autoencoders (MAE) \cite{he2022masked} to train our VIT encoder $\mathcal{T}_e$. This approach enables the ViT encoder to extract relatively universal facial features from facial images with diverse styles. Specifically, we first build a facial image dataset that contains various styles of facial images, such as real-life, cartoon, anime, and, importantly, images from the target engine. We then train the encoder-decoder architecture (see Fig.~\ref{fig:pipeline}(a)) using self-supervised learning on this dataset. Both the encoder and decoder are based on the structure of the vision transformer and we append a global [CLS] token in $\mathcal{T}_e$. During training, each image is divided into $16 \times 16$ patches, and 75\% of these patches are randomly masked before being fed into the ViT encoder. The encoder encodes the masked images into tokens, and the ViT decoder attempts to reconstruct the original image from the encoded tokens using only the pixel-wise L2 loss. Through pre-training on datasets of facial images with various styles, we obtain a universal $\mathcal{T}_e$ that effectively extracts both the structural and cosmetic characteristics of photos.

\subsubsection{Engine-Specific Translator}
By employing the pre-trained $\mathcal{T}_e$ as an image feature extractor, we can further train the translator $\mathcal{T}$ on engine data to output the engine parameters. Specifically, we generate a dataset of parameter-screenshot pairs by randomly sampling the facial structure parameters, makeup texture parameters, and makeup attribute parameters within the avatar customization engine. This dataset is constructed by obtaining rendered screenshots from the customization system corresponding to the sampled parameters.

Based on this specific engine dataset, we train our translator. As shown in Fig.~\ref{fig:pipeline}(b), during the training process, we keep the parameters of the ViT encoder $\mathcal{T}_e$ frozen and train only the parameters of the parameter generation module $\mathcal{T}_p$. The parameter generation module consists of three parallel MLP networks, which generate the facial structure parameters $\delta_s$, makeup texture parameters $\delta_t$, and makeup attribute parameters $\delta_a$, respectively. Each MLP network is composed of a two-layer fully connected (FC) network and takes the [CLS] token as input. For the continuous-valued parameters $\delta_s$ and $\delta_a$, we use the L1 loss. For the discrete-valued parameters $\delta_t$, we employ the cross-entropy loss. Some makeup attribute parameters are only valid under specific textures (e.g., some lipsticks have two layers of color while others have only one). To avoid interference from invalid parameters during loss calculation for these specific makeups, we configure a condition mask (using 0 and 1) to indicate the validity of these makeup attribute parameters.
The total loss function can be formulated as:
\begin{equation}
    \mathcal{L} = \alpha \|\delta_s - \hat{\delta}_s\|_1 + \gamma \|\delta_a \cdot \mathcal{M} - \hat{\delta}_a \cdot \mathcal{M}\|_1 - \lambda \delta_t \cdot \log (\hat{\delta_t}),
\end{equation}
where $\hat{\delta}_s$, $\hat{\delta}_a$, and $\hat{\delta}_t$ represent the predicted values of the facial structure, makeup attribute, and makeup texture parameters, respectively, while $\delta_s$, $\delta_a$, and $\delta_t$ are their corresponding ground truth values. $\mathcal{M}$ denotes the condition mask. The weight coefficients $\alpha$, $\gamma$, and $\lambda$ are set to 5, 1, and 0.1, respectively. During the training process, we also apply commonly used image augmentation techniques to the screenshot images, including random cropping, rotation, color jitter, and Gaussian blur. 

% where $\hat{\delta}_s$, $\hat{\delta}_a$, and $\hat{\delta}_t$ represent the predicted values of the facial structure, makeup attribute, and makeup texture parameters, respectively, while $\delta_s$, $\delta_a$, and $\delta_t$ are their corresponding ground truth values. $\mathcal{M}$ denotes The condition mask. $\alpha$, $\gamma$, and $\lambda$ are weight coefficients and set to 5, 1, and 0.1 respectively. During the training process, we also apply commonly used image augmentation techniques to screenshot images, including random cropping, rotation, color jitter, and Gaussian blur.

Through the second phase of the training process, we  learn the mapping from a unified facial image feature distribution to the specific parameters of an avatar customization system. Although we used only the engine's data for training, the pre-trained general feature extractor $\mathcal{T}_e$ enables our translator to accept images of any style as input and output the corresponding avatar creation parameters specific to the engine. This facilitates photo-based automatic avatar creation. Additionally, since the training in our second phase is conducted solely on randomly sampled paired data from the avatar customization system, our method can be easily extended to other avatar creation systems.

% Through the second phase of the training process, we effectively learn the mapping from a unified facial image feature distribution to specific avatar customization system parameters. Although we used only the Engine's data for training, thanks to the pre-trained general feature extractor $\mathcal{T}_e$, our translator can accept images of any style as input and output the avatar creation parameters specific to the engine, enabling photo-based automatic avatar creation. Additionally, since the training in our second phase is conducted solely on randomly sampled paired data from the avatar customization system, our method can be easily extended to other avatar creation systems.

\begin{figure*}
\centering
\includegraphics[width=1\textwidth]{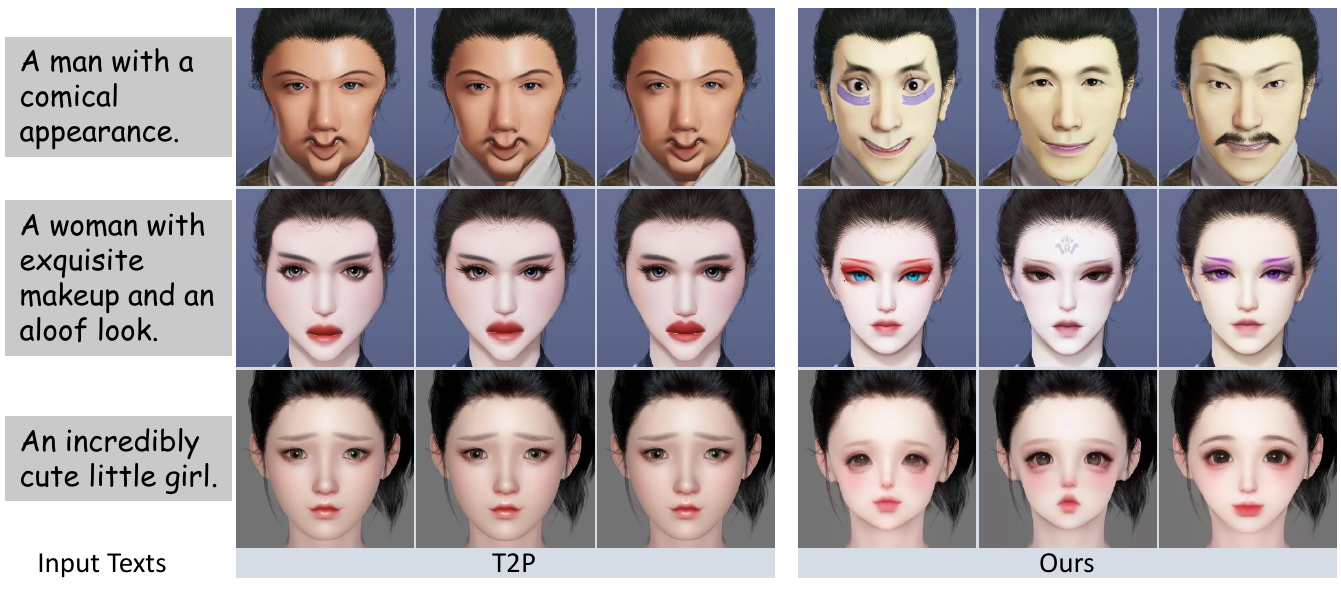}
% {images/ablation_qualitative_head_hightlight.pdf}
\caption{Qualitative comparisons with text-based methods on two avatar customization engines. For each method, we present the results of executing the same text input three times.
}
\label{fig:qualitative_text}
\end{figure*}

\subsection{Text-base Automatic Avatar Creation}\label{text-based}
By using text-to-image techniques such as Stable Diffusion (SD) to generate facial images and employing our translator to convert these images into crafting parameters, we make text-based avatar creation more accessible. However, challenges arise when using the SD model directly. The makeup styles and facial characteristics in images generated by the original SD model often differ significantly from those used in game engines. Although our translator can approximate the avatar customization parameters, this disparity may lead to less accurate results. Furthermore, the inherent unpredictability of the generated images, including occasional failures to consistently produce recognizable facial features, complicates the process and makes it difficult to achieve precise and reliable text-based avatar creation.

% By using text-to-image techniques such as Stable Diffusion (SD) to generate facial images, and subsequently employing our translator to convert these images into crafting parameters, text-based avatar creation becomes more accessible. However, challenges arise when using the SD model directly. The makeup styles and facial characteristics in images generated by the original SD model often differ significantly from those used in game engines. Although our translator can approximate the avatar customization parameters, this disparity may lead to less accurate results. Furthermore, the inherent unpredictability of the generated images, including occasional failures to consistently produce recognizable facial images, further complicates the process, making it difficult to achieve precise and reliable text-based avatar creation. 

To address the aforementioned issues, we train a Stable Diffusion (SD) model to generate avatar facial images consistent with the engine's style. Specifically, we collect a dataset of 7,000 randomly rendered images from the game engine. We then utilize GPT-4o to generate captions describing the makeup, styles, and other facial features of these images. Using this $(\text{image}, \text{caption})$ pair dataset, we fine-tune Stable Diffusion v1.5\footnote{https://huggingface.co/runwayml/stable-diffusion-v1-5}, adjusting both the UNet and text encoder components. Through this fine-tuning process, our model generates facial images that better align with the game engine's visual style, enabling more accurate and reliable text-based avatar creation.

\begin{table*}[] 
\centering
\setlength{\tabcolsep}{1.3mm}{
\begin{tabular}{cccccccc}
\toprule  
&\multicolumn{3}{c}{Justice Mobile}&\multicolumn{3}{c}{Naraka: Bladepoint Mobile} \\
\cmidrule(r){2-4}  \cmidrule(r){5-7} 
Method & Identity Similarity $\uparrow$ & Inception Score$\uparrow$ &  FID$\downarrow$  & Identity Similarity$\uparrow$  & Inception Score$\uparrow$ & FID $\downarrow$ & Speed $\downarrow$  \\
\midrule
F2P~\cite{shi2019face}  &  \textbf{0.376} & \textbf{1.373±0.069} & 40.69 & \textbf{0.334} & \textbf{1.543±0.077} & 42.20 & 1.140s  \\
F2P v2~\cite{shi2020fast}  & 0.275  &  1.134±0.026 & 34.27  & 0.217 &  1.214±0.028  & 33.04 & \textbf{0.007s}\\

% \cmidrule(r){1-8}
Ours  & 0.351 & 1.216±0.043 & \textbf{17.65}  & 0.316 &  1.341±0.059   & \textbf{18.32} & 0.026s\\
\bottomrule 
\end{tabular}
}
\caption{Quantitative comparisons of photo-based methods.}
\label{table:quantitive_image}
\end{table*}

\begin{table*}[!htbp] 
\centering
\setlength{\tabcolsep}{1mm}{
\begin{tabular}{cccccccccc}
\toprule  
&\multicolumn{4}{c}{Justice Mobile}&\multicolumn{4}{c}{Naraka: Bladepoint Mobile} \\
\cmidrule(r){2-5}  \cmidrule(r){6-9} 
Method & LPIPS$\uparrow$ & Inception Score$\uparrow$ &  FID$\downarrow$ & CLIP Scrore $\uparrow$  & LPIPS$\uparrow$  & Inception Score$\uparrow$ & FID $\downarrow$  & CLIP Scrore $\uparrow$ & Speed $\downarrow$\\
\midrule
T2P~\cite{zhao2023zero}     &  0.027 & 1.071±0.007 & 32.9 & 0.211 & 0.014 & 1.098±0.018 & 33.51 & 0.0.223 &  1.725s \\
% \cmidrule(r){1-10}
Ours & \textbf{0.093} & \textbf{1.426±0.104} & \textbf{18.76} & \textbf{0.241} & \textbf{0.095} & \textbf{1.441±0.065} & \textbf{19.43} & \textbf{0.246} & \textbf{0.643s} \\
\bottomrule 
\end{tabular}
}
\caption{Quantitative comparisons of text-based methods.}
\label{table:quantitive_text}
\end{table*}

% \subsection{Implementation Details}

% \textcolor{red}{waiting...}

\section{Experiments}
\subsection{Experimental Data and Implementation Details}
\textbf{Dataset:} To pretrain the ViT encoder, we first construct a large-scale facial image dataset that contains images from existing facial image datasets, including AffectNet~\cite{AffectNet}, CASIA-WebFace~\cite{CASIA-Webface}, CelebA~\cite{CelebA}, IMDB-WIKI~\cite{IMDB-WIKI}, SeepPrettyFace\footnote{www.seeprettyface.com/mydataset.html}, AnimeFace~\cite{AnimeFace}, and Danbooru\footnote{https://huggingface.co/datasets/nyanko7/danbooru2023}. The total number of images is approximately 5.1 million. Our experiments are conducted on two complex avatar customization engines, Justice Mobile and Naraka: Bladepoint Mobile. Each engine features around 400 facial structure parameters, over 200 makeup texture parameters, and more than 100 makeup attribute parameters. For each avatar customization system, we randomly sample 200,000 sets of these parameters and obtain the corresponding rendered images from the engines. These paired parameter-image samples are used to train our translator. Additionally, we incorporate these images into the pretraining of the ViT encoder.

\noindent \textbf{Implementation Details:} All of our models are implemented using PyTorch and are trained with the AdamW~\cite{loshchilov2017decoupled} optimizer. We start by pre-training the ViT encoder on 8 NVIDIA A100 GPUs, using a batch size of 512 and a learning rate of 2e-5 over two weeks on our large dataset. Next, we train the translator with engine data on four NVIDIA A30 GPUs, applying a learning rate of 0.0001 and a batch size of 128 over 50 epochs. For our engine-specific SD model, we employ a batch size of 16 and a learning rate of 5e-6, training on two NVIDIA A100 GPUs for 35 epochs.

\subsection{Comparisons with Photo-Based Methods}
We compare our method with two photo-based automatic avatar creation methods: F2P~\cite{shi2019face} and F2P v2~\cite{shi2020fast}. To facilitate a comprehensive comparison, we collect a test set of 1,000 images encompassing a variety of styles, including real-life, anime, and other two-dimensional art forms. For these images, we first use different methods to obtain the avatar parameters and then generate rendered images through the engine. Performance evaluation is conducted in the image domain by comparing these rendered images. We conduct quantitative assessments using conventional identity similarity metrics~\cite{shi2020fast}, the inception score~\cite{zhao2023zero,salimans2016improved}, and FID (Fréchet Inception Distance)~\cite{wang2023swiftavatar,heusel2017gans}. Specifically, facial similarity is measured using the cosine distance from ArcFace~\cite{deng2019arcface}. For the FID calculation, we select 200 avatars manually created by users and compute the FID distance between these user-created avatars and the ones generated automatically. Additionally, we compare the inference time for each method by presenting the average inference time for each input photo on an RTX 4090.

% We first compare our method with two photo-based automatic avatar creation methods: F2P~\cite{shi2019face} and F2P v2~\cite{shi2020fast}. To facilitate a comprehensive comparison, we collected a test set of 1,000 images encompassing a variety of styles, including real-life, anime, and other two-dimensional art forms. For these images, we first use different methods to obtain the avatar parameters and then generate rendered images through the engine. The performance evaluation is conducted in the image domain by comparing these rendered images. Performance evaluation is conducted in the image domain by comparing these rendered images. We conduct quantitative assessments using conventional identity similarity metrics~\cite{shi2020fast}, the inception score~\cite{zhao2023zero,salimans2016improved}, and FID (Fréchet Inception Distance)~\cite{wang2023swiftavatar,heusel2017gans}. Specifically, facial similarity is measured using the cosine distance from ArcFace~\cite{deng2019arcface}. For the FID calculation, we select 200 avatars manually created by users and compute the FID distance between these user-created avatars and the ones generated automatically. Additionally, we compare the inference time for each method by presenting the average inference time for each input photo on RTX 4090.

The quantitative and qualitative results are presented in Tab.~\ref{table:quantitive_image} and Fig.~\ref{fig:qualitative_image}, respectively. Notably, F2P achieves the highest identity similarity score due to its direct use of identity similarity for supervision. However, this supervision proves inadequate when processing non-realistic photos. As demonstrated in Fig~\ref{fig:qualitative_image}, using non-realistic photos with F2P leads to significantly distorted avatars. Additionally, because the Inception Score mainly assesses the diversity of generated avatars, the numerous unusual results from F2P contribute to its relatively high Inception Score. This diversity deviates considerably from reasonable avatar depictions, as indicated by the large FID score compared to user-created avatars and the visual comparisons in Fig.~\ref{fig:qualitative_image}. In contrast, our method consistently produces reasonable avatars from photos of any style while achieving high identity similarity and the best FID score. Furthermore, our method supports real-time applications on an RTX 4090.

\subsection{Comparisons with Text-Based Methods}
We then compare our method with the state-of-the-art text-based automatic avatar creation method, T2P~\cite{zhao2023zero}. We construct a dataset of 200 textual descriptions of avatar portraits. In our experiments, for each method, we conduct ten inference runs per textual description, resulting in a total of 2,000 evaluation samples for each method. We employ the Learned Perceptual Image Patch Similarity (LPIPS) and Inception Score to assess the diversity of the generated avatars. The LPIPS metric is evaluated on multiple results from the same text prompts. The CLIP score is used to evaluate the semantic consistency between the generated avatars and the input textual descriptions. When computing the CLIP score, the text is uniformly formatted as "a virtual face photo of {}", where {} represents the input text prompt. The Fréchet Inception Distance (FID) score is also utilized to measure the distributional distance between the generated avatars and user-created avatars. Additionally, we compare the inference speed across different methods.

% We then compare our method with the state-of-the-art text-based automatic avatar creation methods, T2P~\cite{zhao2023zero}. We construct a dataset of 200 textual descriptions of avatar portraits. In our experiments, for each method, we conduct ten inference runs per textual description, resulting in a total of 2000 evaluation samples for each method. We employ the Learned Perceptual Image Patch Similarity (LPIPS) and Inception Score to assess the diversity of the generated avatars. The LPIPS metric is evaluated on multiple results from the same text prompts. The CLIP score is used to evaluate the semantic consistency between the generated avatars and the input textual descriptions. When computing the CLIP score, the text is uniformly formatted as ``a virtual face photo of {}'', where {} represents the input text prompt. The Fréchet Inception Distance (FID) score is also utilized to measure the distributional distance between the generated avatars and user-created avatars. Additionally, we also compare the inference speed across different methods.

Tab.~\ref{table:quantitive_text} and Fig.~\ref{fig:qualitative_text} demonstrate the quantitative and qualitative results, respectively. It is evident that our method surpasses the T2P approach across all metrics, showcasing the effectiveness and robustness of our approach. As seen in Fig.~\ref{fig:qualitative_text}, compared to T2P, our method generates more diverse and accurate results for the same textual input. These comparative results validate the superiority of our method.

% Tab.~\ref{table:quantitive_text} and Fig.~\ref{fig:qualitative_text} demonstrate the quantitative and qualitative results, respectively. It is evident that our method surpasses the T2P approach across all metrics, showcasing the effectiveness and robustness of our approach. As can be seen from Fig.~\ref{fig:qualitative_text}, compared to T2P, our method is capable of generating more diverse and accurate results for the same textual input. These comparative results validate the superiority of our method.

\begin{figure}
\centering
\includegraphics[width=0.46\textwidth]{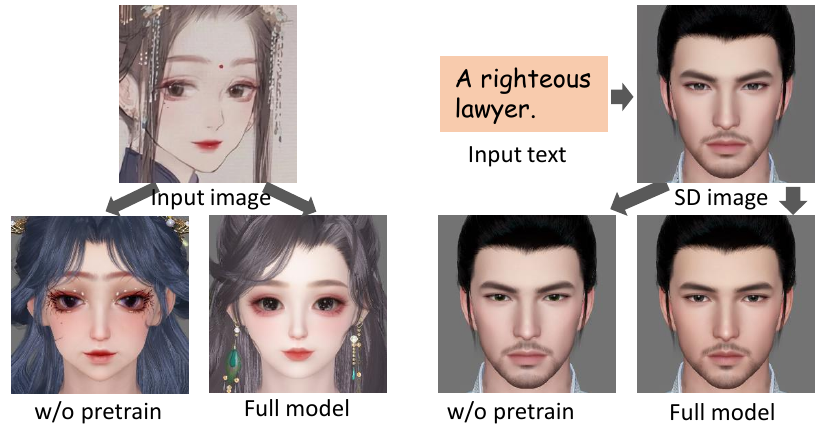}
\caption{Qualitative evaluations without pretraining of the VIT encoder. The bottom row depicts the generated avatars.
}
\label{fig:ablation_vit}
\end{figure}

\begin{figure}
\centering
\includegraphics[width=0.44\textwidth]{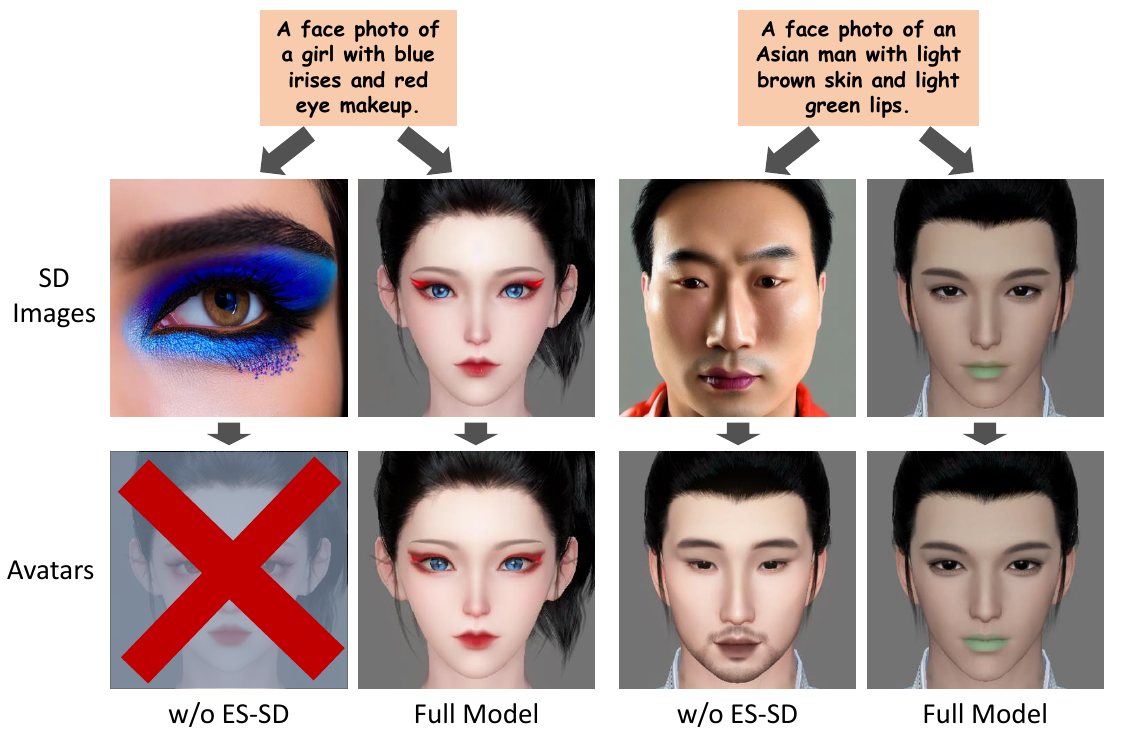}
\caption{Ablation study of removing our engine-specific SD. For w/o ES-SD, we use the original SD with the text prompt above to generate images, and then apply our Translator to convert the images into avatar parameters.
}
\label{fig:ablation_sd}
\end{figure}

\subsection{Ablation Study}
To evaluate the key components of our proposed method, we conduct two ablation study experiments on Justice Mobile. We first remove the pretraining of the ViT encoder $\mathcal{T}_e$ (w/o pretrain) and train all parameters of the translator $\mathcal{T}$ on engine data. Tab.~\ref{table:ablation_study_image} and Tab.~\ref{table:ablation_study_text} show the quantitative evaluations of photo-based avatar auto-creation and text-based avatar auto-creation, respectively, while Fig.~\ref{fig:ablation_vit} presents the qualitative evaluations. From Tab.~\ref{table:ablation_study_image} and Fig.~\ref{fig:ablation_vit}, it is evident that removing the pretraining of the ViT encoder results in a significant deterioration of both metrics and visual results. Although the Inception Score increases substantially, this is due to the generation of numerous distorted results, leading to an apparent but unreasonable diversity. On the other hand, as seen in Tab.~\ref{table:ablation_study_text} and Fig.~\ref{fig:ablation_vit}, the impact of ViT encoder pretraining on text-based avatar auto-creation is minimal. This is because the stable diffusion of our engine's style ensures that the generated images adhere to a specific style. Even without ViT pretraining, our translator can accurately convert images into avatar parameters. This demonstrates the flexibility of our method; if image-based avatar creation is not required, ViT pretraining can be entirely skipped.

% To evaluate the key components of proposed method, we conduct two ablations study experiments on Justice Mobile. We first remove the pretrain of ViT encoder $\mathcal{T}_e$ (w/o pretrain) and training all parameters of the translator $\mathcal{T}$ on engine data. Tab.~\ref{table:ablation_study_image} and Tab.~\ref{table:ablation_study_text} shows the quantitative evaluations on photo-based avatar auto-creation and text-avatar auto-creation, respectively, and Fig.~\ref{fig:ablation_vit} shows the qualitative evaluations. From Tab.~\ref{table:ablation_study_image} and Fig.~\ref{fig:ablation_vit}, it is evident that once the pretraining of the VIT encoder is removed, both the metrics and results deteriorate significantly. Although the Inception Score increases substantially, this is due to the generation of numerous distorted results, leading to an apparent but unreasonable diversity. On the other hand, as seen from Tab.~\ref{table:ablation_study_text} and Fig.~\ref{fig:ablation_vit}, the impact of VIT encoder pretraining on text-based avatar auto-creation is minimal. This is because our engine style's stable diffusion ensures that the generated images adhere to a specific style. Even without VIT pretraining, our translator can accurately convert images into avatar parameters. This demonstrates the flexibility of our method, if image-based avatar creation is not required, VIT pretraining can be entirely skipped.

We then conduct another ablation study by replacing the engine-style SD with the original SD (w/o ES-SD). Fig.~\ref{fig:ablation_sd} shows the evaluation results. As observed, although we can directly use the original SD to generate images, we find that even with very strict prompt control, SD cannot always ensure the generation of images with fully frontal faces. Additionally, SD itself lacks a strong semantic understanding of makeup details, making it prone to errors when handling specific facial makeup features. In contrast, our engine-style SD enhances the understanding of makeup details and consistently generates images in the engine style, achieving more accurate and robust text-based avatar creation.

% We then perform another ablation study by replacing the engine style SD with the original SD (w/o ES-SD). Fig.~\ref{fig:ablation_sd} shows the evaluation results. As observed, although we can directly use SD to generate images, we find that even with very strict prompt control, SD cannot always guarantee generating images with full frontal faces. Additionally, SD itself does not have a strong semantic understanding of makeup details, making it prone to errors when dealing with facial makeup specifics. In contrast, our engine style SD enhances the understanding of makeup details and ensures that images in the engine style are consistently generated, achieving more accurate and robust text-based avatar creation.

% We then perform another ablation study by replacing the engine style SD with the origin SD (w/o ES-SD). Fig.~\ref{fig:ablation_sd} shows the evaluation results.  

\subsection{User Study}

\begin{table}
\small
\centering
\setlength{\tabcolsep}{1mm}{
\begin{tabular}{cccc}
\toprule  
Method & Identity Similarity $\uparrow$ & Inception Score$\uparrow$ &  FID$\downarrow$ \\
\midrule
w/o pretrain & 0.243 & \textbf{2.060± 0.109} & 97.31
 \\
Full model & \textbf{0.351} & 1.216±0.043 & \textbf{17.65} \\
\bottomrule 
\end{tabular}}
\caption{Quantitative results of the ablation study on photo-based avatar auto-creation.}
\label{table:ablation_study_image}
\end{table}

\begin{table}
\small
\centering
\setlength{\tabcolsep}{1mm}{
\begin{tabular}{ccccc}
\toprule  
Method & LPIPS$\uparrow$ & Inception Score$\uparrow$ &  FID$\downarrow$ & CLIP Scrore $\downarrow$  \\
\midrule
w/o pretrain & 0.092 & \textbf{1.428± 0.105} & \textbf{18.73} & \textbf{0.241}
 \\
Full model & \textbf{0.093} & 1.426±0.104 & 18.76 & \textbf{0.241} \\
\bottomrule 
\end{tabular}}
\caption{Quantitative results of the ablation study on text-based avatar auto-creation.}
\label{table:ablation_study_text}
\end{table}

\begin{table}
% \small

\centering
\setlength{\tabcolsep}{2mm}{
\begin{tabular}{ccc}
\toprule  
Method & Fidelity Score (1-5) &  Prefer Ratio \\
\midrule
F2P~\cite{shi2019face} & 2.24 & 11\% \\
F2P v2~\cite{shi2020fast} & 1.67 & 2\% \\
ours & \textbf{3.97} & \textbf{87\%} \\
\bottomrule 
\end{tabular}}
\caption{Results of the user study on photo-based methods.}
\label{table:user_study_image}
\end{table}

\begin{table}
% \small

\centering
\setlength{\tabcolsep}{2mm}{
\begin{tabular}{ccc}
\toprule  
Method & Consistency Score (1-5) &  Prefer Ratio \\
\midrule
T2P~\cite{zhao2023zero}  & 2.74 & 23\% \\
ours & \textbf{4.41} & \textbf{77\%} \\
\bottomrule 
\end{tabular}}
\caption{Results of the user study on Text-based methods.}
\label{table:user_study_text}
\end{table}

We also conduct two user study groups involving 50 participants to further demonstrate the effectiveness and robustness of our method. For the photo-based methods, we randomly select 100 sets of test results for each participant from a total of 2,000 sets (1,000 test images on two engines). Each set includes the input image and the rendered avatars generated by our method, F2P, and F2P v2. These are presented to the participants, who are asked to rate the fidelity of each result compared to the original image on a scale from 1 to 5, and to select the one they consider the best. For the text-based methods, we randomly select 100 sets of test results for each participant from a total of 2,000 sets (200 textual descriptions with 5 generations each on two engines). Each set includes the input text and the rendered avatars generated by our method and T2P. These are presented to participants, who are asked to rate the consistency of each result with the textual description and to choose the avatar they prefer.  Tab.~\ref{table:user_study_image} and Tab.~\ref{table:user_study_text} show the results of our user study. As can be seen, our method significantly outperforms other methods across various user study metrics, demonstrating the superiority of our approach.

% We also conduct two groups of user study involving 50 participants to further demonstrate the effectiveness and robustness of our method.  For the photo-based methods, we randomly select 100 sets of test results for each participant from a total of 2000 sets (1000 test images on two engines). Each set includes the input image and the rendered avatars generated by our method, F2P, and F2P v2. These are presented to the participants, who are asked to rate the fidelity of each result compared to the original image on a score of 1 to 5, and to select the one they consider the best. The results are shown in Tab.~\ref{table:user_study_image}. 

% For the text-based methods, we randomly select 100 sets of test results for each participant from a total of 2000 sets (200 textual descriptions with 5 generations each on two engines). Each set includes the input text and the rendered avatars generated by our method and T2P. These are presented to the participants, who are asked to rate the consistency of each result with the textual description and to choose the avatar they prefer. The results are shown in Tab.~\ref{table:user_study_text}. 
% From Tab.~\ref{table:user_study_image} and Tab.~\ref{table:user_study_text}, it is evident that our method significantly outperforms other methods across various user study metrics, demonstrating the superiority of our approach.

\section{Conclusion}
In this paper, we introduce EasyCraft, a unified framework that integrates photo-based and text-based avatar auto-creation through a Translator that converts images into avatar parameters. By employing self-supervised pretraining on the image feature encoder within the Translator, our method supports images of any style as input, eliminating the previous reliance on off-the-shelf supervision. Our Translator relies on training solely with game engine data, which allows our approach to be more easily applied to other avatar customization systems. Our proposed method departs from prior frameworks based on imitator and inversion techniques, providing new insights into automatic engine-based parametric avatar creation.

% In this paper, we introduce EasyCraft, which integrates photo-based avatar auto-creation and text-based avatar auto-creation into a unified framework through the use of a Translator that converts images into avatar parameters. By employing self-supervised pretraining on the image feature encoder within the Translator, our method can support images of any style as input, thus eliminating the previous reliance on off-the-shelf supervision. Our translator relies training solely on the game engine, this flexibility allows our approach to be more easily applied to other avatar customization systems. Our proposed method departs from prior frameworks based on imitator and inversion techniques, providing new insights into automatic engine-based parametric avatar creation.

{
    \small
    \bibliographystyle{ieeenat_fullname}
    \bibliography{main}
}


\end{document}

%% file: preamble.tex
\usepackage[dvipsnames]{xcolor}